\PassOptionsToPackage{table,xcdraw}{xcolor}
\documentclass[11pt]{mystyle}

\usepackage[numbers,sort&compress]{natbib}
\usepackage{nicefrac}       %
\usepackage{ragged2e}
\usepackage{multirow}
\usepackage{makecell}
\usepackage{hhline}
\usepackage{subcaption}
\usepackage{adjustbox}
\usepackage{float}
\usepackage{placeins}
\usepackage{listings}
\usepackage{needspace}
\tcbuselibrary{listings,breakable,skins}

\definecolor{promptaccent}{HTML}{405D78}
\definecolor{promptback}{HTML}{FBFCFD}
\definecolor{promptframe}{HTML}{C9D1D9}
\definecolor{prompttitle}{HTML}{F1F4F7}
\definecolor{prompttext}{HTML}{24292F}
\newtcblisting{promptbox}[1]{
  enhanced,
  breakable,
  listing only,
  title={#1},
  colback=promptback,
  colframe=promptframe,
  colbacktitle=prompttitle,
  coltitle=promptaccent,
  fonttitle=\bfseries\small,
  boxrule=0.35pt,
  titlerule=0.35pt,
  borderline west={1.4pt}{0pt}{promptaccent},
  arc=1pt,
  outer arc=1pt,
  left=1.6mm,
  right=1.6mm,
  top=1.2mm,
  bottom=1.2mm,
  before skip=5pt,
  after skip=9pt,
  listing options={
    basicstyle=\ttfamily\footnotesize\color{prompttext},
    columns=fullflexible,
    keepspaces=true,
    showstringspaces=false,
    breaklines=true,
    breakatwhitespace=true,
    postbreak={}
  }
}

\newcommand\blfootnote[1]{\begingroup\renewcommand\thefootnote{}\footnotetext{#1}\endgroup}

\tcbset{titlebox/.append style={left=6mm,right=6mm}}
\title{Can Large Language Model Agents Learn\\
by Watching Tutorial Videos?}

\DeclareRobustCommand{\paperauthorblock}{%
  {\normalfont\normalsize\bfseries
  Haoyu Zhao \quad Zafeirios Fountas \quad Jun Wang$^{\dagger}$\par}
  \vspace{4pt}
  {\normalfont\fontsize{10}{12}\selectfont
  \makebox[\linewidth][c]{University College London}\par}
  \vspace{2pt}
}
\author{\paperauthorblock}
\date{}

\begin{document}

\begin{abstract}
Humans continuously acquire knowledge from carefully crafted materials such as books and tutorial videos, carrying that understanding into future tasks. Millions of such instructional videos are uploaded online every day, yet LLM agents remain unable to exploit this ever-expanding wealth of human teaching. Their capabilities are frozen after training, while parameter updates at scale remain prohibitively expensive. This raises the central question: can LLM agents learn from this vast instructional resource without parameter updates? We introduce Tutorial Agentic Learning (TAL), a nonparametric continual-learning paradigm that consolidates tutorial videos into a persistent, structured tutorialbook memory through a two-stage adversarial process that harvests hard, video-grounded questions and uses them to refine the memory. At inference time, TAL retrieves through a Table-of-Contents (TOC) tree, tying retrieval directly to the model's reasoning rather than to embedding similarity. To study this setting, we introduce VidLearn, a multimodal QA benchmark built from 72 chess tutorial videos spanning 12 openings. On VidLearn, TAL outperforms all evaluated baselines and improves monotonically as more videos are consumed, achieving up to 113\% and 91\% relative gains over the corresponding vanilla MLLMs on the training and test splits, respectively. In actual gameplay, TAL reaches a stronger opening-phase position than its vanilla counterpart in 69\% of head-to-head games. Evaluation in a second domain further demonstrates the generality of TAL. Together, these results highlight instructional videos as a scalable resource for parameter-update-free agent continual learning.

\end{abstract}
\maketitle
\blfootnote{$^{\dagger}$Corresponding author: \texttt{jun.wang@cs.ucl.ac.uk}}
\thispagestyle{plain}
\pagestyle{plain}

\section{Introduction}
Large language models (LLMs) and LLM-based agents have demonstrated impressive capabilities across a wide range of domains~\cite{anthropic2025claude,openai2025gpt5,xai2025grok4,zhao2025seecontrol}. Yet under conventional parametric learning, their capabilities are largely fixed once training is complete, and enabling such systems to improve continually after deployment remains a central challenge~\cite{fang2025comprehensive,gao2025survey}. Updating these models typically requires training signals from human-labeled data or environmental feedback~\cite{turing2004intelligent,silver2025welcome}, yet repeated fine-tuning is prohibitively expensive at scale. Humans, by contrast, routinely learn from instructional materials such as books and tutorial videos~\cite{bandura1986social,mayer2005cambridge}. Millions of new instructional videos are uploaded online every day~\cite{brame2017effective}, creating a rapidly growing multimodal resource that today's LLM agents cannot yet fully leverage. This raises the central question: can LLM agents learn from this vast instructional resource without parameter updates?

\begin{figure}[t]
  \centering
  \includegraphics[width=\linewidth]{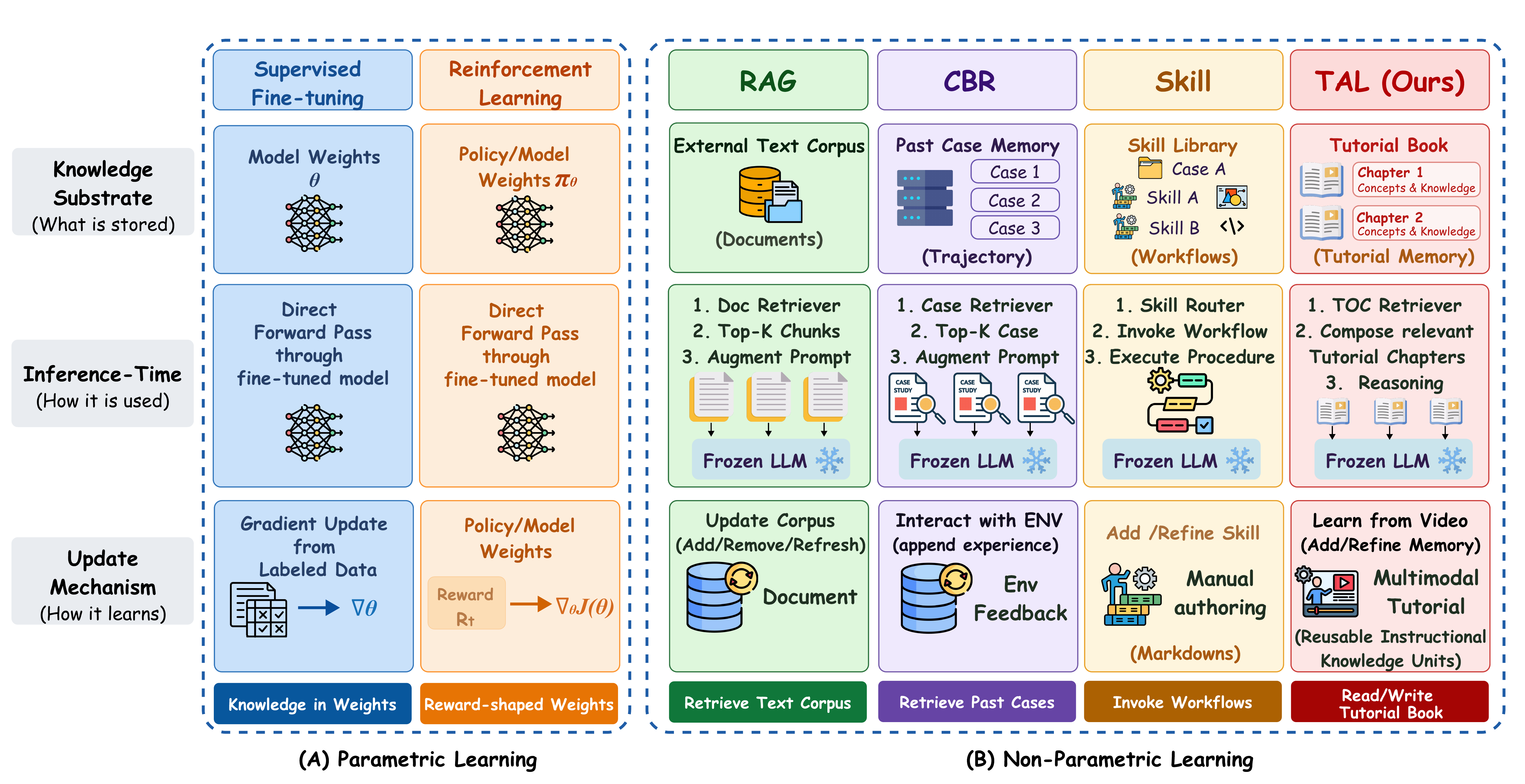}
  \caption{Overview of learning paradigms for LLM agents, comparing parametric and non-parametric approaches along the dimensions of stored knowledge, inference-time usage, and update mechanism.}
  \label{fig:tal_com}
\end{figure}

One promising direction is the recent line of nonparametric mechanisms for LLM agents~\cite{fang2025comprehensive,gao2025survey, silver2025welcome}, in which capability is grown through external memory rather than continual weight updates. As summarized in Fig.~\ref{fig:tal_com}, this line splits into two forms. In the first, knowledge is supplied externally by humans, as in retrieval-augmented generation (RAG)~\cite{lewis2020retrieval} and agent skills~\cite{anthropic2026skills}. RAG retrieves textual evidence from human-curated corpora, while agent skills inject human-authored workflows specified in structured formats such as markdown. These approaches are effective, but they rely on knowledge manually curated for agent consumption and therefore do not scale with the open-world supply of human teaching. In the second form, knowledge is derived from the model's own past interactions, as in case-based reasoning (CBR)~\cite{watson1994case}, where adaptation proceeds by retrieving similar prior cases. Because such memory records what happened rather than why or how, CBR offers limited interpretive guidance in novel contexts, and its effectiveness depends heavily on case coverage and the model's ability to reason from raw trajectories.

These limitations motivate a third direction in which agents learn reusable knowledge from tutorial videos. Unlike curated RAG corpora and hand-crafted skills, tutorial videos already exist at internet scale and require no agent-specific curation; unlike case-based experience, they do not merely record what happened, but pair demonstrations with explanations of \emph{why} and \emph{how}, providing direct interpretive guidance for future reasoning. Yet today's LLM agents lack a mechanism to convert raw tutorial videos into persistent, reusable knowledge. Videos are information-rich, inherently multimodal, making this conversion nontrivial. If such structured knowledge could be consolidated from tutorial videos and organized for reuse, an agent could improve from human teaching without modifying its parameters.

To realize this, we introduce \textbf{Tutorial Agentic Learning (TAL)}, a nonparametric continual-learning paradigm that consolidates tutorial videos into a structured tutorial memory---a ``tutorialbook''---that a frozen LLM can read at inference time. TAL learns this memory through a two-stage consolidation. \emph{Stage I} harvests hard, video-grounded questions: a video agent generates question--answer pairs supported by evidence, and only those the frozen model fails to answer without external context are retained. \emph{Stage II} turns these failures into targeted memory edits via a read--write loop: TAL revisits the relevant tutorial context, validates the missing knowledge exposed by each failure, and rewrites the corresponding chapters to incorporate the missing knowledge. At inference time, TAL retrieves through a Table-of-Contents (TOC) tree: the model first scans the chapter titles to identify relevant chapters, then expands those chapters to access the underlying details. This ties retrieval to the model's reasoning rather than to embedding similarity, eliminating the mismatch between semantic similarity and task relevance. TAL generalizes beyond both RAG-style evidence lookup and case-level replay by turning human teaching videos into an external knowledge structure that can be incrementally refined and reused.

To study this setting, we construct \textbf{VidLearn}, a multimodal QA benchmark built from 72 chess tutorial videos spanning 12 classical openings. Chess offers a controlled yet realistic testbed by combining structured knowledge, an abundant supply of online tutorials, rich visual--language grounding, and nontrivial multi-step reasoning. On VidLearn, TAL outperforms strong fine-tuning-free baselines by up to 113\% on the training split and 91\% on the test split, and the learned memory transfers to actual play, where TAL reaches a stronger opening-phase position than the vanilla baseline in 69\% of head-to-head games, as evaluated by Stockfish. Crucially, TAL's performance scales monotonically with the number of consumed tutorial videos, demonstrating that frozen LLM agents can continuously accumulate reusable knowledge from instructional videos without parameter updates. To assess cross-domain generalization, we additionally evaluate TAL in a software-tutorial setting. Across matched comparisons with stronger retrieval and memory baselines, TAL achieves the highest mean accuracy in all eight chess settings and reaches 49.7\% held-out accuracy in the software domain, 10.9 percentage points above the strongest baseline.

Our main contributions are:
\begin{itemize}
\item We introduce Tutorial Agentic Learning (TAL), a nonparametric continual-learning paradigm that enables frozen LLM agents to learn from instructional videos without parameter updates.
\item We design a two-stage adversarial consolidation that harvests hard, video-grounded questions and uses them to refine a structured tutorial memory, together with a TOC-based retrieval mechanism that ties inference-time retrieval to the model's own reasoning.
\item We introduce VidLearn, a benchmark of 72 chess tutorial videos across 12 openings, and provide extensive evaluations on VidLearn and a separate software-tutorial setting, demonstrating TAL's effectiveness, scalability, and cross-domain generalization.
\end{itemize}

\section{Related Work}
\subsection{Continual Learning for LLM Agents}
A growing body of work enables LLM agents to improve from experience without fine-tuning, relying on external memory rather than parameter updates~\cite{zhao2024expel,tang2025agent,chhikara2025mem0,wang2024agent,ouyang2025reasoningbank,cai2025flex,zhao2026cle}. These efforts fall into two paradigms. \emph{Skill-based} methods~\cite{anthropic2026skills,wang2023voyager,jiang2026xskill} store reusable workflows for on-demand invocation, but constructing high-quality workflows often requires manual curation, limiting their scalability to online instructional resources. \emph{Experience-based} methods, rooted in case-based reasoning~\cite{watson1994case,guo2024ds} and exemplified by recent systems such as Memento~\cite{Memento-II,zhao2026memento3}, store and retrieve prior agent-generated cases through reflective read--write loops. Related reflection-based methods, such as Reflexion~\cite{shinn2023reflexion} and Self-Refine~\cite{madaan2023self}, use failures or feedback to iteratively improve individual trajectories or outputs. While effective, case-level reuse depends on the availability of similar past interactions and on the model's ability to reason from retrieved cases. TAL departs from both by consolidating knowledge from human tutorial videos into a structured, reusable knowledge substrate.
\subsection{Learning from Demonstrations}
Learning from human demonstrations has long been a central theme in robotics~\cite{argall2009survey,atkeson1997robot,ravichandar2020recent,schaal1996learning}, where agents learn by imitating expert trajectories. However, demonstrations in these settings are typically distilled into policies or action sequences for task execution and control~\cite{brohan2022rt,zitkovich2023rt} rather than into reusable knowledge structures. More recently, multimodal LLM research has targeted long-form single-video   understanding~\cite{zhou2025mlvu,Lvbench,Longvideobench}, where models struggle as videos grow longer and denser. Retrieval-augmented approaches such as VideoRAG~\cite{luo2024video,ren2025videorag,jeong2025videorag} and in-context methods such as Demo-ICL~\cite{dong2026icl} address this by retrieving relevant segments before generation. Yet these works primarily target inference-time comprehension of individual videos, whereas TAL converts tutorial videos into structured external knowledge that frozen LLM agents can reuse across future tasks.

\section{Tutorial Agentic Learning}
We study continual learning from a stream of instructional videos $\mathcal{D}=\{v_t\}_{t=1}^{T}$, where $v_t$ is a tutorial video accompanied by $N_t$ downstream multimodal QA pairs $\{(x_{t,i},a_{t,i}^\star)\}_{i=1}^{N_t}$. The base solver is a frozen model $f_\theta$ with parameters $\theta$, i.e., $\theta_{t+1}=\theta_t=\theta$ throughout the learning stream, so adaptation cannot rely on parameter updates. The central problem is therefore how to acquire reusable knowledge by watching tutorial videos without changing $\theta$. This setting is motivated by how humans learn from instructional media: they watch tutorials to acquire knowledge and later consult structured tutorialbooks when solving new problems.

\begin{figure}[t]
  \centering
  \includegraphics[width=\linewidth]{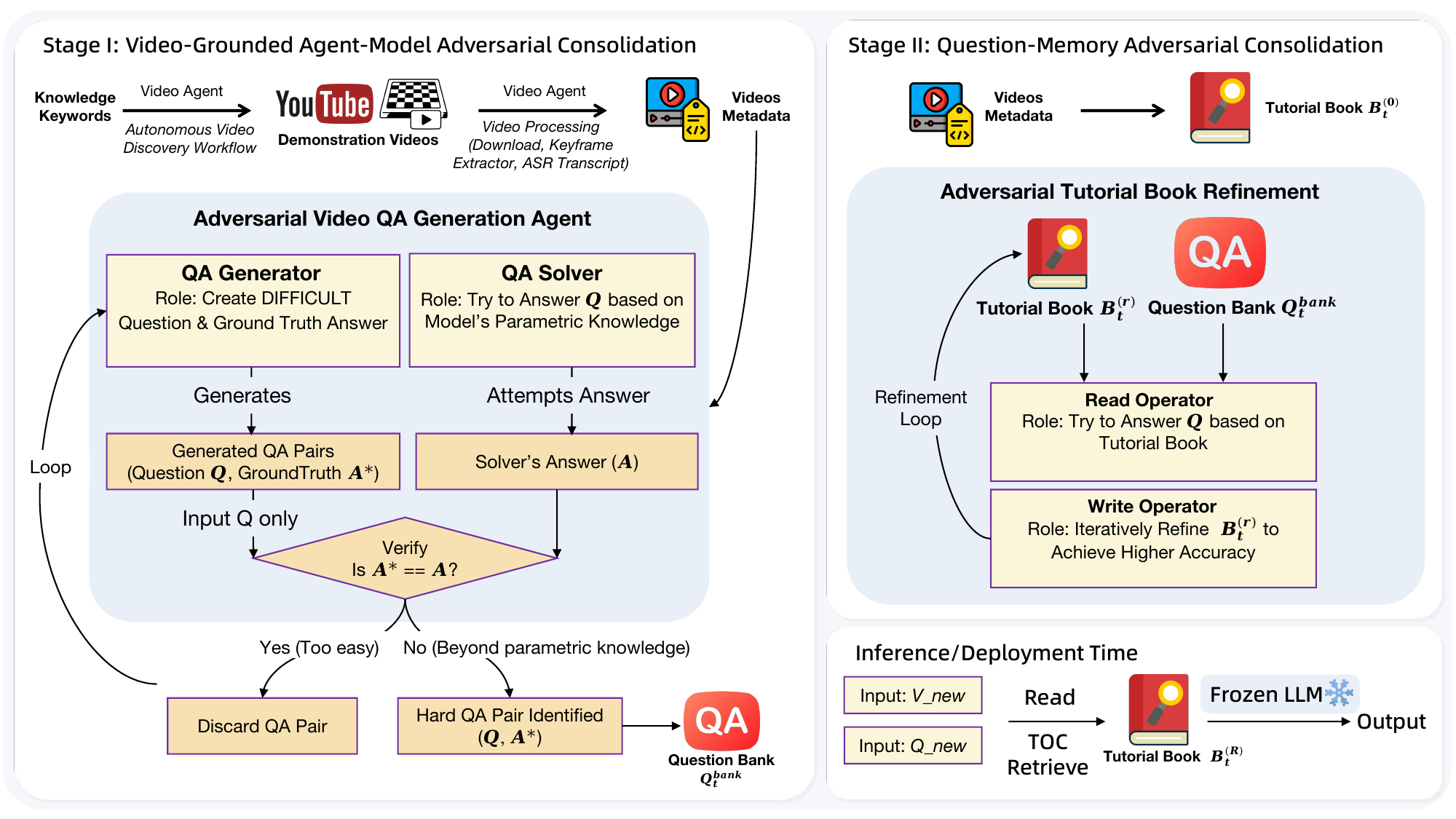}
  \caption{Overview of Tutorial Agentic Learning (TAL). In Stage I, TAL performs video-grounded agent--model adversarial consolidation, where a video agent harvests hard question--answer pairs from instructional videos that are supported by tutorial evidence yet unsolved by the frozen model, forming a question bank. In Stage II, TAL performs question--memory adversarial consolidation, using this question bank to iteratively refine tutorial memory through a read--write loop. At inference time, TAL retrieves relevant tutorial units via a Table-of-Contents (TOC) tree, enabling the frozen model to access structured tutorial knowledge for downstream reasoning without parameter updates.}
  \label{fig:tal_overview}
\end{figure}

\subsection{Tutorialbook Memory as a Learning Substrate}
We contrast three learning substrates:
(1) \emph{Parametric learning}, where knowledge is implicitly stored in model parameters and adaptation is achieved through weight updates.
(2) \emph{Case-level replay}, where adaptation is achieved by retrieving stored episodic cases, typically produced by the agent's own interactions, while keeping the model frozen.
(3) \emph{TAL tutorial memory}, also referred to as a \emph{tutorialbook}, where adaptation is achieved by writing and reading structured knowledge consolidated from multimodal human tutorials without changing model weights. TAL therefore shifts the learning substrate from parameter space or raw episodic traces to an evolving tutorialbook memory that consolidates multimodal tutorial content with its supporting evidence into a reusable tutorialbook:
\begin{equation}
B_t^{(r)}=\{(k_j,z_j^{(r)})\}_{j=1}^{N_t^{(r)}},
\end{equation}
  where $r \in \{0,1,\ldots,R\}$ denotes the refinement round, $R$ is the total number of rounds, $B_t^{(R)}$ is the final refined memory, $k_j$ is the chapter's retrieval key (category, title, and short summary), and $z_j^{(r)}$ is a textbook-style chapter with an overview, thematic sections (each elaborated by bullet points and key terms), and a summary that consolidates the takeaways.

\subsection{Two-Stage Adversarial Consolidation}
TAL learns tutorial memory through a two-stage adversarial consolidation: Stage I harvests hard, video-grounded questions exposing the frozen model's gaps, and Stage II turns these failures into targeted memory edits via a read--write loop \cite{Memento-II}. We detail each stage below.

\subsubsection{Stage I: Video-Grounded Agent--Model Adversarial Consolidation}
\label{sec:stage1}
Stage I constructs the hard question bank used to supervise tutorial-memory refinement in Stage II. To do so, TAL instantiates a video-grounded agent--model adversarial process over each video. A video agent analyzes aligned visual and textual evidence to generate candidate questions, while the frozen solver attempts to answer them without external context. The goal is to retain questions that are both supported by the tutorial evidence and difficult for the frozen model.

For each tutorial video, the video agent parses the video and subtitles, aligns multimodal evidence by timestamp, and samples an evidence snippet $e=(\tau, W_\tau, I_\tau)$ from the video. Concretely, each snippet consists of (i) a timestamp anchor $\tau$ chosen by the planner from the subtitle timeline, (ii) the local subtitle window $W_\tau$ spanning a $\pm 10$\,s neighborhood of $\tau$, and (iii) the video frame $I_\tau$ extracted at $\tau$; together they pin a single teaching moment to a specific (text, vision) pair. The agent then generates a grounded question $q \sim G_\phi(v_t,e)$ via a question generator $G_\phi$ (where $\phi$ is updated each round through reflective revision of easy-question failures), derives a reference answer $a_q^\star$ from $e$, and forms a multimodal query $x_q$ pairing $q$ with $I_\tau$. We measure the difficulty of a candidate question under an empty external context $c_0$ as
\begin{equation}
d_t(q)=\ell\!\left(f_\theta(x_q,c_0),a_q^\star\right),\qquad
\rho_t(q)=\mathbf{1}\!\left[(q,a_q^\star)\text{ is supported by }v_t\right],
\end{equation}
where $\ell(\cdot,\cdot)$ denotes the multiple-choice prediction loss with respect to the reference answer. Only hard and grounded questions are retained:
\begin{equation}
\delta_t(q)=\mathbf{1}\!\left[d_t(q)>\epsilon_d\right]\rho_t(q).
\end{equation}
The retained items form the Stage I question bank
\begin{equation}
\mathcal{Q}_t^{\text{bank}}=\{(x_q,a_q^\star):\delta_t(q)=1\}.
\end{equation}

In practice, $\rho_t(q)$ is instantiated as a \emph{grounding gate}: an independent LLM judge inspects the local subtitle window $W_\tau$ together with $(q, a_q^\star)$ and admits the candidate only if the question can be directly grounded in the lecturer's content at $\tau$, rather than recoverable from general domain knowledge or other parts of the video. The judge also rejects perception-only questions. Since both $q$ and $a_q^\star$ are derived from $e$, the gate also constrains $a_q^\star$ to be evidence-supported, preventing the agent from grading its own parametric guesses. The difficulty signal $d_t(q)$ is then obtained by querying the frozen solver on $x_q$ with empty external context $c_0$ and scoring its prediction against $a_q^\star$. These QA pairs drive memory refinement in Stage II.

\subsubsection{Stage II: Question--Memory Adversarial Consolidation}
\label{sec:stage2}
Starting from an initial memory $B_t^{(0)}$ constructed from the structured multimodal evidence snippets associated with the harvested questions, TAL repeatedly tests the current memory against the question bank. Questions that are incorrectly answered expose knowledge that is missing, incomplete, or insufficiently organized in the current memory; these failures then guide targeted memory edits.

At refinement round $r$, TAL uses the current tutorial memory to answer each query in $\mathcal{Q}_t^{\text{bank}}$:
\begin{equation}
\hat{a}_q^{(r)}=
\arg\max_a p_\theta(a\mid x_q,\mathrm{Read}(x_q,B_t^{(r)})).
\end{equation}
Here $p_\theta(a\mid x_q,\mathrm{Read}(x_q,B_t^{(r)}))$ denotes the frozen model's answer distribution under the retrieved tutorial context, and $\arg\max$ selects the predicted answer $\hat{a}_q^{(r)}$. The failure set then consists of the queries that remain incorrect under the current memory:
\begin{equation}
\mathcal{F}_t^{(r)}=
\left\{(x_q,a_q^\star)\in\mathcal{Q}_t^{\text{bank}}:\hat{a}_q^{(r)}\neq a_q^\star\right\}.
\end{equation}
A write operator then updates tutorial memory by proposing edits to chapters implicated in unresolved failures and accepting them only if they improve performance on
  $\mathcal{Q}_t^{\text{bank}}$:
\begin{equation}
B_t^{(r+1)}=
\operatorname{Write}\!\left(B_t^{(r)},\mathcal{Q}_t^{\text{bank}},\mathcal{F}_t^{(r)}\right),
\end{equation}
where $\mathcal{F}_t^{(r)}$ provides edit targets and $\mathcal{Q}_t^{\text{bank}}$ serves as a scoring set. Concretely, the Write operator first proposes a candidate via an LLM-based editor that rewrites the chapters relevant to each failure. The candidate is then re-scored on $\mathcal{Q}_t^{\text{bank}}$ and accepted only if it strictly improves over $B_t^{(r)}$; otherwise $B_t^{(r)}$ is retained. This propose--and--verify procedure guarantees monotonic improvement across rounds and prevents destabilizing previously correct knowledge. The loop runs for at most $R$ iterations, with early stopping when no candidate improves accuracy. We denote the resulting memory as $B_t^{(R)}$.

\subsubsection{Learning in Tutorial Memory}
In TAL, learning takes place in tutorial memory rather than in model parameters. The central objective is to consolidate each tutorial video $v_t$ into a memory $\mathcal{M}_t$ (realized in TAL as the tutorialbook $B_t^{(R)}$) that is simultaneously compact and sufficient for downstream queries. We motivate the design of TAL by viewing it through the lens of the Information Bottleneck (IB) principle~\cite{tishby2000information}:
\begin{equation}
\label{eq:ib}
\min_{\mathcal{M}_t}\; I(\mathcal{M}_t;\,V_t)\;-\;\beta\,I(\mathcal{M}_t;\,Y_t),
\end{equation}
where compressing $I(\mathcal{M}_t;\,V_t)$ prevents the memory from becoming an unselective encoding of the full multimodal video content, and maximizing $I(\mathcal{M}_t;\,Y_t)$ ensures it remains predictive of downstream answers. TAL does not optimize this objective directly---doing so would require estimating mutual information over multimodal distributions, which is intractable. Instead, TAL operationalizes the same tradeoff through a tractable two-stage procedure that can be viewed as a sequential minimax structure~\cite{fan1953minimax}: Stage I updates the question generator's memory $\phi$ via reflective write to maximize the harvesting objective
\begin{equation}
\mathcal{J}^{\mathrm{harvest}}_t(\phi)=
\mathbb{E}_{e\sim E(v_t),\,q\sim G_\phi(v_t,e)}
\left[d_t(q)\rho_t(q)\right],
\end{equation}
where $d_t(q)$ measures how difficult question $q$ is for the frozen model under empty context, and $\rho_t(q)$ keeps only questions that are supported by tutorial evidence. Given the harvested supervision set, Stage II updates tutorial memory $B_t^{(r)}$ via the Write operator to minimize the memory-conditioned answer loss
\begin{equation}
\label{eq:mem_loss}
\mathcal{L}^{\mathrm{mem}}_t\!\left(B_t^{(r)}\right)=
\frac{1}{|\mathcal{Q}_t^{\text{bank}}|}
\sum_{(x_q,a_q^\star)\in\mathcal{Q}_t^{\text{bank}}}
\ell\!\left(f_\theta(x_q,\mathrm{Read}(x_q,B_t^{(r)})),a_q^\star\right)
\;+\;
\gamma\,\mathcal{C}\!\left(B_t^{(r)},B_t^{(r-1)}\right).
\end{equation}

Here $\mathcal{C}(B_t^{(r)},B_t^{(r-1)})$ is a proximity penalty that keeps consecutive memory states close, formalizing the localized-edit constraint introduced in Sec.~\ref{sec:stage2}: in practice, the Write operator combines an LLM-based rewrite with a score-based acceptance test: the LLM is prompted to edit chapters implicated by the current failure set, and the resulting candidate $B_t^{(r+1)}$ is accepted only if its accuracy on $\mathcal{Q}_t^{\text{bank}}$ strictly improves over $B_t^{(r)}$; otherwise the candidate is discarded and $B_t^{(r)}$ is retained. This guarantees monotonic improvement and prevents destabilizing previously correct knowledge. This update rule echoes the structure of proximal optimization methods such as mirror descent~\cite{beck2003mirror}, where each accepted step is constrained to the neighborhood of the previous iterate. Together, the harvesting and memory-conditioned terms define the coupled minimax objective
\begin{equation}
\min_{B_t^{(R)}}\max_{\phi}
\underbrace{\mathcal{J}^{\mathrm{harvest}}_t(\phi)}_{\text{hard-question harvesting}}
\;+\;
\lambda
\underbrace{\mathcal{L}^{\mathrm{mem}}_t(B_t^{(R)})}_{\text{memory-conditioned answer loss}},
\end{equation}
where $B_t^{(r)}$ is updated at each round via the Write operator and $\phi$ via reflective memory revision, neither requiring gradient descent. The entire learning loop is therefore fully nonparametric---one stage refining question-generation memory, the other refining tutorial memory, both without any parameter updates.

\subsection{Inference with Tutorial Memory}
After tutorial memory has been refined, TAL answers new queries by retrieving and composing relevant tutorial knowledge. We formalize inference as a memory-conditioned answer distribution
\begin{equation}
\pi(a\mid x, B_t^{(R)})
= \sum_{c \in B_t^{(R)}} \mu(c \mid x, B_t^{(R)})\, p_\theta(a \mid x, c),
\end{equation}
where $\mu(\cdot \mid x, B_t^{(R)})$ is a retrieval distribution over tutorial chapters and $p_\theta(a \mid x, c)$ is the frozen answer kernel conditioned on retrieved context $c$. In practice, $\mu$ is realized as a deterministic top-$k$ selector via TOC retrieval, so the mixture collapses to a single retrieved context
$\mathrm{Read}(x, B_t^{(R)})$, and the final prediction reduces to
\begin{equation}
\hat{y} = \arg\max_a \pi(a \mid x, B_t^{(R)})
= f_\theta\!\left(x,\, \mathrm{Read}(x, B_t^{(R)})\right).
\end{equation}
\subsection{TOC Retrieval}
  \label{sec:toc}
TAL instantiates $\mathrm{Read}(x, B_t^{(R)})$ through a Table-of-Contents (TOC) retrieval mechanism mirroring how humans use a book: skim the index, decide which chapters to read, then expand only those chapters in detail. The tutorial memory is structured as a hierarchical TOC tree of \emph{volumes}, \emph{parts}, and \emph{chapters} (the atomic content units). At inference, the frozen LLM reads the TOC entries with the query $x$, selects the top-$k$ chapter IDs $\mathcal{C}_x$, and these chapters are aggregated into the retrieved context
\begin{equation}
\mathrm{Read}(x, B_t^{(R)}) = \mathrm{aggregate}\bigl(\{z_i^{(R)} : i \in \mathcal{C}_x\}\bigr),
\end{equation}
where $z_i^{(R)}$ is the full content of chapter $i$. Retrieval is thus driven by the LLM's reasoning rather than embedding similarity, integrating retrieval into the model's chain of reasoning.
\section{Experiment}
\subsection{Dataset}
We use chess as a controlled yet realistic testbed for learning from instructional videos. Chess offers four advantages for this purpose: (i) a well-codified body of teachable knowledge developed over centuries~\cite{feng2023chessgpt}; (ii) an abundant supply of tutorial videos on YouTube, providing scalable supervision; (iii) multimodal instruction that pairs visual board states with natural-language explanations; and (iv) substantial difficulty for current LLMs~\cite{wen2025chessqa,kolasani2025llm}, leaving room for measurable progress.

Based on these properties, we introduce VidLearn, a benchmark of 72 tutorial videos spanning 12 classical openings (Appendix~\ref{sec:vidlearn_details}). For each opening, we collect six videos via YouTube search using the opening name as the query. The benchmark contains 48 training videos and 110 multimodal test questions derived from 24 held-out videos, with all test annotations manually verified. During training, the agent is given only the YouTube links. At test time, it answers each question using only a single video frame, a natural-language query, and multiple-choice options.
\subsection{Experimental Setup}
We use two complementary evaluation protocols. First, to assess robustness across model families, we compare four methods: (i) vanilla direct prompting, (ii) prompting with thinking mode enabled, (iii) Transcript-RAG, and (iv) TAL. All four methods are evaluated across four different model families, including GPT-5.2~\cite{openai2025gpt52}, Kimi K2.5~\cite{team2026kimi}, Gemini 3 Flash~\cite{deepmind2025gemini3flash}, and Claude Sonnet 4.6~\cite{anthropic2026sonnet46}. Second, to compare TAL with stronger retrieval and memory systems under a matched setting, we fix the backbone to Gemini-3.5-Flash~\cite{deepmind2026gemini35flash} at temperature $0.7$ and evaluate vanilla prompting, Transcript-RAG, two Video-RAG variants~\cite{luo2024video}, mem0~\cite{chhikara2025mem0}, ReasoningBank~\cite{ouyang2025reasoningbank}, and TAL. Full implementation details are provided in Appendix~\ref{sec:baseline_experimental_configuration}.

\begin{samepage}
In Stage I, each video undergoes up to three rounds of adversarial question harvesting. Each round segments the tutorial timeline into 20 event candidates, each anchored to a timestamped frame and its surrounding subtitles. It then generates one question per event. We cap harvesting at 60 questions per video to ensure broad temporal coverage while avoiding diminishing returns from long-tail mining. In Stage II, memory refinement runs for at most three iterations, with early stopping based on validation performance. At inference, TOC retrieval selects the top three chapters as tutorial evidence. For consistency, all cross-model TAL experiments use the same GPT-5.2-generated tutorialbook, allowing us to evaluate whether the learned tutorial memory transfers across models.
\par\end{samepage}

\begin{table}[t]
\centering
\caption{Comparison of baseline methods and TAL on the training and test sets under varying numbers of demonstration videos. Best results are in \textbf{bold}. TAL consistently outperforms all baselines across all settings, showing a clear scaling trend as more tutorial demonstrations become available.}
\label{tab:exp_main}
\small
\renewcommand{\arraystretch}{0.7}
\resizebox{\linewidth}{!}{%
\begin{tabular}{lcccccccc}
\toprule
Method & \multicolumn{4}{c}{Training Set} & \multicolumn{4}{c}{Test Set} \\
\cmidrule(lr){2-5}\cmidrule(lr){6-9}
\# Demonstration Videos & 12 & 24 & 36 & 48 & 12 & 24 & 36 & 48 \\
\midrule
\multicolumn{9}{l}{\textbf{Vanilla MLLMs~(Direct Answering)}} \\
\midrule
GPT-5.2 & 31.61\% & 31.61\% & 31.61\% & 31.61\% & 37.27\% & 37.27\% & 37.27\% & 37.27\% \\
Kimi K2.5  & 38.51\% & 38.51\% & 38.51\% & 38.51\% & 31.82\% & 31.82\% & 31.82\% & 31.82\% \\
Gemini 3 Flash & 47.70\% & 47.70\% & 47.70\% & 47.70\% & 48.18\% & 48.18\% & 48.18\% & 48.18\% \\
Claude Sonnet 4.6 & 41.95\% & 41.95\% & 41.95\% & 41.95\% & 37.27\% & 37.27\% & 37.27\% & 37.27\% \\
\midrule
\multicolumn{9}{l}{\textbf{MLLMs~+~CoT (Reasoning)}} \\
\midrule
GPT-5.2 & 29.31\% & 29.31\% & 29.31\% & 29.31\% & 30.91\% & 30.91\% & 30.91\% & 30.91\% \\
Kimi K2.5  & 35.63\% & 35.63\% & 35.63\% & 35.63\% & 32.73\% & 32.73\% & 32.73\% & 32.73\% \\
Gemini 3 Flash & 45.98\% & 45.98\% & 45.98\% & 45.98\% & 44.55\% & 44.55\% & 44.55\% & 44.55\% \\
Claude Sonnet 4.6 & 41.95\% & 41.95\% & 41.95\% & 41.95\% & 34.55\% & 34.55\% & 34.55\% & 34.55\% \\
\midrule
\multicolumn{9}{l}{\textbf{MLLMs~+~RAG (Embedding~+~BM25 Hybrid Document Retrieval)}} \\
\midrule
GPT-5.2  & 34.48\% & 43.68\% & 46.55\% & 51.15\% & 42.73\% & 38.18\% & 43.64\% & 37.27\% \\
Kimi K2.5 & 47.13\% & 47.70\% & 48.85\% & 50.57\% & 40.91\% & 48.18\% & 38.18\% & 35.45\% \\
Gemini 3 Flash  & 53.45\% & 54.02\% & 56.90\% & 63.79\% & 52.73\% & 52.73\% & 50.91\% & 49.09\% \\
Claude Sonnet 4.6 & 43.68\% & 54.02\% & 53.45\% & 54.60\% & 38.18\% & 40.00\% & 41.82\% & 41.82\% \\
\midrule
\multicolumn{9}{l}{\textbf{Ours~(MLLMs~+~TAL)}} \\
\midrule
\rowcolor{blue!8}
TAL (GPT-5.2) & \textbf{43.68\%} & \textbf{51.15\%} & \textbf{58.62\%} & \textbf{67.24\%} & \textbf{48.18\%} & \textbf{57.27\%} & \textbf{61.82\%} & \textbf{64.55\%} \\
\rowcolor{blue!8}
TAL (Kimi K2.5) & \textbf{48.28\%} & \textbf{54.02\%} & \textbf{58.62\%} & \textbf{64.94\%} & \textbf{46.36\%} & \textbf{52.73\%} & \textbf{59.09\%} & \textbf{60.91\%} \\
\rowcolor{blue!8}
TAL (Gemini 3 Flash) & \textbf{64.94\%} & \textbf{62.64\%} & \textbf{80.46\%} & \textbf{81.61\%} & \textbf{54.55\%} & \textbf{57.27\%} & \textbf{69.09\%} & \textbf{71.82\%} \\
\rowcolor{blue!8}
TAL (Claude Sonnet 4.6) & \textbf{56.90\%} & \textbf{55.75\%} & \textbf{63.22\%} & \textbf{71.84\%} & \textbf{39.09\%} & \textbf{42.73\%} & \textbf{49.09\%} & \textbf{54.55\%} \\
\bottomrule
\end{tabular}
}
\end{table}
\subsection{Main Results}
During Stage~I, TAL harvests 174 multimodal QA instances from the 48 training videos, which serve as the question bank $\mathcal{Q}_t^{\text{bank}}$ for Stage~II memory refinement. We report accuracy on these harvested questions to measure how much of the targeted knowledge TAL successfully absorbs into memory, and on the held-out test set to measure generalization to unseen videos. Table~\ref{tab:exp_main} summarizes the main results. Overall, TAL achieves the best performance on both splits and exhibits the strongest scaling behavior with respect to the number of tutorial videos. Relative to the corresponding vanilla MLLM baselines, TAL attains a maximum relative gain of 112.72\% on the training set (GPT-5.2) and 91.42\% on the test set (Kimi K2.5), showing that frozen LLM agents can substantially benefit from instructional videos without any parameter updates.

A key pattern in Table~\ref{tab:exp_main} is the contrast between TAL and the other fine-tuning-free baselines as the training corpus grows. RAG improves training performance by exposing the model to subtitle evidence, but its gains on the test set are inconsistent and often deteriorate at larger corpus sizes. By contrast, TAL continues to improve as more tutorial videos are added. This suggests that retrieval over raw subtitle chunks becomes increasingly noisy with scale, making it harder to surface evidence that is genuinely useful for reasoning. Enabling the model's built-in thinking mode also fails to yield consistent improvements over direct answering, and in some cases performs slightly worse. Taken together, these results indicate that the main bottleneck in this setting is not simply inference-time reasoning depth, but access to the right instructional knowledge. TAL specifically addresses this bottleneck via structured tutorial memory and TOC retrieval.

\begin{figure}[!htbp]
  \centering
\includegraphics[width=\linewidth]{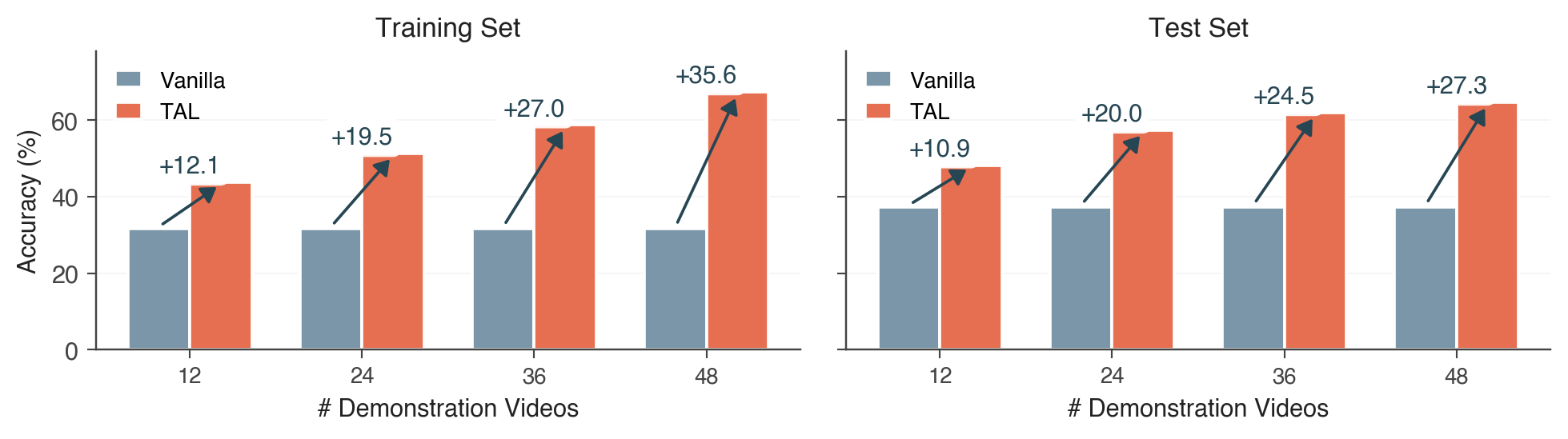}
  \caption{Scaling behavior of TAL vs.\ vanilla (GPT-5.2) on training (left) and test (right) splits.}
  \label{fig:tal_vs_vanilla_test_test}
\end{figure}

\textbf{Ability Scaling.} TAL leads at every video budget, and its advantage widens monotonically with corpus size (Fig.~\ref{fig:tal_vs_vanilla_test_test}): each additional tutorial is consolidated into progressively richer memory with broader topical coverage, so that TAL accumulates reusable knowledge over time rather than merely caching individual demonstrations. The resulting trajectory mirrors a human-like learning curve, for which we provide a reference under the same setting in Appendix~\ref{sec:human_performance}. Beyond accuracy, TAL also yields a much more compact knowledge representation: with 48 training videos, the raw subtitle corpus used by RAG contains 598{,}027 tokens, whereas the resulting tutorial memory contains only 105{,}703 tokens---an 82.32\% reduction, while still delivering higher accuracy on both splits.

\textbf{Cross-Model Transferability.} Beyond scaling, TAL improves performance across all four model families, indicating that its gains are not tied to any single answer model. The tutorial memory built by GPT-5.2 can be consumed directly by other LLMs at inference via TOC retrieval, without rebuilding, suggesting that TAL externalizes instructional knowledge in a model-agnostic form that can be built once and reused across answer models.

\paragraph{Comparison with Retrieval and Memory Baselines.}
We further compare TAL against two Video-RAG variants~\cite{luo2024video} and two recent text-based memory systems, mem0~\cite{chhikara2025mem0} and ReasoningBank~\cite{ouyang2025reasoningbank}. All methods use the same Gemini-3.5-Flash backbone at temperature $0.7$, with results reported as mean$\pm$standard deviation over three seeds. Implementation details are provided in Appendix~\ref{sec:baseline_experimental_configuration}.

\begin{table}[htbp]
\centering
\caption{Comparison with stronger retrieval and memory baselines. Best results are in \textbf{bold}.}
\label{tab:strong_baselines_chess}
\small
\setlength{\tabcolsep}{4pt}
\resizebox{\linewidth}{!}{%
\begin{tabular}{lcccccccc}
\toprule
Method & \multicolumn{4}{c}{Training Set (174 QA)} & \multicolumn{4}{c}{Test Set (110 QA)} \\
\cmidrule(lr){2-5}\cmidrule(lr){6-9}
\# Demonstration Videos & 12 & 24 & 36 & 48 & 12 & 24 & 36 & 48 \\
\midrule
Vanilla (no memory) & \multicolumn{4}{c}{53.1$\pm$1.0} & \multicolumn{4}{c}{57.6$\pm$2.3} \\
\midrule
Transcript-RAG & 55.9$\pm$0.5 & 58.2$\pm$1.1 & 60.0$\pm$2.3 & 64.6$\pm$3.1 & 59.7$\pm$1.9 & 59.4$\pm$0.4 & 59.7$\pm$1.5 & 60.0$\pm$2.6 \\
Video-RAG (caption retrieval)~\cite{luo2024video} & 58.0$\pm$0.9 & 59.6$\pm$0.3 & 59.4$\pm$2.7 & 62.1$\pm$1.7 & 58.5$\pm$0.9 & 57.9$\pm$0.9 & 62.1$\pm$1.5 & 64.5$\pm$1.5 \\
Video-RAG (+retrieved frames)~\cite{luo2024video} & 59.2$\pm$2.0 & 57.3$\pm$2.2 & 60.0$\pm$2.7 & 62.1$\pm$2.6 & 60.3$\pm$1.5 & 61.8$\pm$3.7 & 62.4$\pm$2.3 & 63.6$\pm$0.7 \\
mem0~\cite{chhikara2025mem0} & 57.1$\pm$0.5 & 55.7$\pm$1.4 & 58.2$\pm$1.0 & 59.6$\pm$0.7 & 62.7$\pm$2.0 & 63.0$\pm$1.7 & 62.4$\pm$0.4 & 60.9$\pm$1.5 \\
ReasoningBank~\cite{ouyang2025reasoningbank} & 56.5$\pm$1.4 & 55.6$\pm$0.7 & 61.1$\pm$0.7 & 59.6$\pm$1.8 & 58.5$\pm$2.3 & 59.7$\pm$2.4 & 61.2$\pm$2.4 & 60.6$\pm$3.7 \\
\rowcolor{blue!8}
\textbf{TAL (ours)} & \textbf{68.2$\pm$1.2} & \textbf{70.7$\pm$1.7} & \textbf{77.8$\pm$1.2} & \textbf{84.1$\pm$1.4} & \textbf{64.5$\pm$2.0} & \textbf{64.8$\pm$1.1} & \textbf{66.1$\pm$0.4} & \textbf{71.8$\pm$2.0} \\
\bottomrule
\end{tabular}%
}
\end{table}

As shown in Table~\ref{tab:strong_baselines_chess}, TAL achieves the highest mean accuracy in all eight training/test settings. With 48 videos, it reaches 84.1\% on the training questions and 71.8\% on the held-out set, compared with 64.6\% and 64.5\% for the strongest respective baselines.

\paragraph{Generalization Beyond Chess.}
\label{sec:second_domain_evaluation}
TAL's question harvesting, memory refinement, and TOC retrieval do not rely on chess-specific representations.
To assess cross-domain generalization, we apply TAL to software tutorials covering four Adobe Photoshop 2026 features: Generative Fill, Harmonize, Object Removal \& Selection, and AI Adjustments \& Workspace. Following the same evaluation protocol as in the main VidLearn experiments, we collect 28 videos, using 16 for learning and holding out 12 for evaluation, and construct 108 QA pairs comprising 53 training questions and 55 held-out test questions.

We compare TAL with Vanilla, Transcript-RAG, two Video-RAG variants~\cite{luo2024video}, mem0~\cite{chhikara2025mem0}, and ReasoningBank~\cite{ouyang2025reasoningbank}. All methods use Gemini-3.5-Flash at temperature $0.7$, and we report mean$\pm$standard deviation over three seeds.

\begin{table}[htbp]
\centering
\caption{Second-domain evaluation on Adobe Photoshop 2026 software tutorials. Best results are in \textbf{bold}.}
\label{tab:photoshop_second_domain}
\small
\setlength{\tabcolsep}{4pt}
\resizebox{\linewidth}{!}{%
\begin{tabular}{lcccccccc}
\toprule
Method & \multicolumn{4}{c}{Training Set (53 QA)} & \multicolumn{4}{c}{Test Set (55 QA)} \\
\cmidrule(lr){2-5}\cmidrule(lr){6-9}
\# Demonstration Videos & 4 & 8 & 12 & 16 & 4 & 8 & 12 & 16 \\
\midrule
Vanilla (no memory) & \multicolumn{4}{c}{38.4$\pm$5.4} & \multicolumn{4}{c}{21.2$\pm$3.7} \\
\midrule
Transcript-RAG & 62.3$\pm$2.7 & 66.7$\pm$0.9 & 72.3$\pm$0.9 & 78.6$\pm$0.9 & 27.3$\pm$2.6 & 35.8$\pm$0.9 & 29.7$\pm$3.1 & 30.9$\pm$2.6 \\
Video-RAG (caption retrieval)~\cite{luo2024video} & 68.6$\pm$0.9 & 72.3$\pm$2.4 & 81.8$\pm$0.9 & 84.3$\pm$3.2 & \textbf{33.9$\pm$0.9} & 38.2$\pm$8.3 & 29.7$\pm$3.1 & 38.8$\pm$4.8 \\
Video-RAG (+retrieved frames)~\cite{luo2024video} & 67.3$\pm$1.8 & 74.8$\pm$3.2 & 84.3$\pm$0.9 & 86.2$\pm$1.8 & 32.1$\pm$0.9 & 36.4$\pm$1.5 & 27.3$\pm$1.5 & 33.9$\pm$2.3 \\
mem0~\cite{chhikara2025mem0} & 54.1$\pm$3.6 & 57.2$\pm$0.9 & 67.9$\pm$1.5 & 71.7$\pm$1.5 & 29.7$\pm$5.6 & 33.9$\pm$1.7 & 33.9$\pm$4.8 & 33.9$\pm$1.7 \\
ReasoningBank~\cite{ouyang2025reasoningbank} & 46.5$\pm$3.2 & 45.3$\pm$2.7 & 45.9$\pm$6.4 & 49.7$\pm$1.8 & 24.2$\pm$3.1 & 24.8$\pm$5.6 & 23.0$\pm$4.5 & 27.9$\pm$2.3 \\
\rowcolor{blue!8}
\textbf{TAL (ours)} & \textbf{69.2$\pm$3.9} & \textbf{82.4$\pm$0.9} & \textbf{96.2$\pm$1.5} & \textbf{98.7$\pm$0.9} & 32.1$\pm$0.8 & \textbf{40.0$\pm$1.5} & \textbf{35.8$\pm$3.7} & \textbf{49.7$\pm$5.2} \\
\bottomrule
\end{tabular}%
}
\end{table}

As the training corpus grows from 4 to 16 videos, TAL's training accuracy increases from 69.2\% to 98.7\% (Table~\ref{tab:photoshop_second_domain}). With all 16 training videos, TAL reaches 49.7\% on held-out questions, 10.9 percentage points above the strongest baseline. TAL obtains the highest mean in all four training settings and in three of the four held-out settings, demonstrating TAL's effectiveness across domains.

\vspace{-6pt}
\subsection{Head-to-Head Evaluation}
To evaluate whether TAL improves practical move-level decision quality in direct play, we construct a controlled head-to-head chess environment. In this setting, both agents use the same base model (GPT-5.2): a \textit{vanilla} agent that plays from model priors alone, and a \textit{TAL} agent that plays with tutorial-memory retrieval enabled. We evaluate all 12 openings covered by VidLearn. For each opening, games begin from a fixed opening-specific position. To reduce side bias, we adopt a color-swapped protocol and run 20 games per opening (10 each as White and Black).

We assess opening-phase quality at move horizons $N\in\{2,3,4,5\}$ using two metrics: (1) Win rate: at the move-$N$ cutoff, the side with the higher Stockfish evaluation is credited with a win (draws count as $0.5$), and we aggregate win rates across all games. This metric captures which agent obtains the more favorable position after a fixed number of moves. (2) Average CentiPawn Loss (ACPL): we compute the per-move deviation of each side's played move from the Stockfish-optimal move (one pawn $=100$ centipawns); lower values indicate higher move quality. As shown in Table~\ref{tab:head2head_acpl_winrate}, TAL achieves higher win rates and lower ACPL than the vanilla baseline at every horizon, indicating that tutorial memory translates into measurably better move-level decisions in practical play.

\begin{table*}[th]
\centering
\caption{Head-to-head TAL vs. vanilla results under fixed openings.}
\label{tab:head2head_acpl_winrate}
\resizebox{\textwidth}{!}{%
\begin{tabular}{lcccc}
\toprule
Horizon & Win Rate $\uparrow$ (TAL/Van) & Mean ACPL $\downarrow$ (TAL/Van) & Median ACPL $\downarrow$ (TAL/Van) & 5\% Trimmed Mean ACPL $\downarrow$ (TAL/Van) \\
\midrule
move2 & \textbf{50.83\%} / 49.17\% & \textbf{63.14} / 85.26 & \textbf{11.50} / 24.50 & \textbf{47.98} / 74.36 \\
move3 & \textbf{69.17\%} / 30.83\% & \textbf{87.62} / 142.98 & \textbf{28.83} / 135.17 & \textbf{75.01} / 130.75 \\
move4 & \textbf{67.92\%} / 32.08\% & \textbf{107.34} / 153.51 & \textbf{66.25} / 146.75 & \textbf{99.96} / 145.38 \\
move5 & \textbf{62.50\%} / 37.50\% & \textbf{125.74} / 162.81 & \textbf{112.60} / 156.60 & \textbf{119.24} / 159.01 \\
\bottomrule
\end{tabular}
}
\end{table*}

\subsection{Ablation Study}
\subsubsection{Effect of Two-Stage Consolidation}
To isolate the contribution of each stage, we evaluate two reduced variants. Stage I only uses the harvested QA pairs directly as case memory at inference time, reducing TAL to a case-based reasoning (CBR)-style system. Stage II only retains the initial tutorialbook constructed from video but disables Stage II refinement against the harvested question bank. Both stages contribute substantially (Table~\ref{tab:ablation_stage12}): Stage I only improves over vanilla from 48.18\% to 55.45\%, while Stage II only reaches a similar 56.36\%. Full TAL reaches 71.82\%, 8.2 points above the additive prediction of 63.63\%, suggesting that the two stages interact synergistically.

\subsubsection{Effect of Retrieval Strategy}
We isolate the retrieval mechanism via a 12-way opening classification: a strategy is correct if its top-3 retrieved chapters include the query's source opening. With the tutorialbook and base model fixed, TOC retrieval achieves $50.00\%$ Recall@3, compared with $28.18\%$ for dense retrieval using OpenAI's text-embedding-3-small (Table~\ref{tab:ablation_retrieval_backend}). This 21.82-point improvement indicates that reasoning over the tutorialbook structure better captures opening-level relevance than embedding similarity.

\begin{table}[H]
\centering
\begin{minipage}[t]{0.60\linewidth}
\vspace{0pt}
\centering
\captionof{table}{Two-stage consolidation ablation.}
\label{tab:ablation_stage12}
\footnotesize
\renewcommand{\arraystretch}{1.05}
\begin{adjustbox}{max width=\linewidth}
\begin{tabular}{@{}lccc@{}}
\toprule
Setting & Stage I & Stage II & Test Score \\
\midrule
Vanilla Setting & No & No & 48.18\% \\
Stage I only (w/ CBR, w/o Tutorialbook) & Yes & No & 55.45\% \\
Stage II only (w/ Tutorialbook, w/o Refinement) & No & Yes & 56.36\% \\
Stage I + Stage II (Full TAL) & Yes & Yes & \textbf{71.82\%} \\
\bottomrule
\end{tabular}
\end{adjustbox}
\end{minipage}\hfill
\begin{minipage}[t]{0.37\linewidth}
\vspace{0pt}
\centering
\captionof{table}{Retrieval-strategy ablation.}
\label{tab:ablation_retrieval_backend}
\footnotesize
\renewcommand{\arraystretch}{1.05}
\begin{adjustbox}{max width=\linewidth}
\begin{tabular}{@{}lc@{}}
\toprule
Setting & Recall@3 \\
\midrule
TAL + Embedding Retrieval & 28.18\% \\
TAL + TOC Retrieval & \textbf{50.00\%} \\
\bottomrule
\end{tabular}
\end{adjustbox}
\end{minipage}
\end{table}

\begingroup
\setlength{\intextsep}{7pt plus 2pt minus 1pt}
\captionsetup[table]{skip=5pt}
\subsubsection{Effect of Visual Information}
To isolate whether visual information contributes during tutorial-memory construction, we compare two otherwise matched TAL variants on the same 110 held-out questions. Full TAL constructs the tutorialbook from event-aligned video frames and transcripts, whereas transcript-only TAL removes all visual inputs and constructs the tutorialbook solely from subtitle text. We use a Gemini-3.5-Flash answer model at temperature $0.7$ and report the mean and standard deviation over three seeds.

\begin{table}[htbp]
  \centering
  \caption{Vision-in-construction ablation on 110 held-out questions. Full TAL uses event-aligned frames and transcripts, while transcript-only TAL removes all visual inputs. Results are mean$\pm$standard deviation over three seeds; best results are in \textbf{bold}.}
  \label{tab:vision_construction_ablation}
  \small
  \setlength{\tabcolsep}{5pt}
  \begin{adjustbox}{max width=\linewidth}
  \begin{tabular}{lcccc}
    \toprule
    Memory built from & \multicolumn{4}{c}{Test accuracy (\%) at \# demonstration videos} \\
    \cmidrule(lr){2-5}
     & 12 & 24 & 36 & 48 \\
    \midrule
    Vanilla (no memory) & \multicolumn{4}{c}{57.6$\pm$2.3} \\
    Transcript-only TAL (no visual input) & 58.8$\pm$1.4 & 58.5$\pm$2.3 & 60.9$\pm$5.5 & 62.1$\pm$1.9 \\
    \rowcolor{blue!8}
    Full TAL (frames + transcripts) & \textbf{64.5$\pm$2.0} & \textbf{64.8$\pm$1.1} & \textbf{66.1$\pm$0.4} & \textbf{71.8$\pm$2.0} \\
    \bottomrule
  \end{tabular}
  \end{adjustbox}
\end{table}

Full TAL outperforms transcript-only TAL at every video budget, reaching 71.8\% versus 62.1\% with 48 videos. During memory construction and refinement, event-aligned frame--subtitle pairs connect the lecturer's explanation to concrete board states, pieces, squares, and spatial relations. Because the TAL pipeline is otherwise held fixed, this comparison directly isolates the contribution of visual evidence rather than differences in retrieval or memory design. This consistent advantage shows that TAL leverages visual information during learning, supporting the complementary value of tutorial videos beyond their transcripts.

\titlespacing*{\subsubsection}{0pc}{1ex plus1pt minus1pt}{0pt}
\subsubsection{Impact of Retrieval Depth}
Table~\ref{tab:ablation_topk} reveals a crossover between $k=3$ and $k=5$ as the tutorial corpus grows. With sparse memory (12--24 videos), $k=5$ outperforms $k=3$ by $2.7$--$3.6$ points, as additional retrieved chapters help compensate for the limited per-query coverage of an immature tutorialbook. Once memory matures past 24 videos, the ranking flips: at 48 videos, $k=3$ reaches $71.82\%$ while $k=5$ reaches only $66.36\%$. We attribute this to a coverage--dilution tradeoff: when individual chapters are already information-dense, surfacing additional ones increasingly introduces marginally relevant content rather than new evidence.

\begin{table}[htbp]
\centering
\caption{Impact of retrieval depth.}
\label{tab:ablation_topk}
\small
\begin{tabular}{lcccc}
\toprule
Setting & \multicolumn{4}{c}{\# Demonstration Videos} \\
\cmidrule(lr){2-5}
Top-$k$ & 12 & 24 & 36 & 48 \\
\midrule
$k=3$ & 54.55\% & 57.27\% & 69.09\% & 71.82\% \\
$k=5$ & 58.18\% & 60.00\% & 60.91\% & 66.36\% \\
\bottomrule
\end{tabular}
\end{table}

\endgroup

\titlespacing*{\section}{0pc}{4pt}{1pt}
\section{Conclusion}
\titlespacing*{\section}{0pc}{3ex plus4pt minus3pt}{5pt}
We introduced Tutorial Agentic Learning (TAL), a nonparametric continual-learning paradigm that enables frozen LLM agents to consolidate instructional videos into a structured tutorialbook memory accessed through TOC retrieval. Across VidLearn and a second software-tutorial evaluation, TAL outperforms representative fine-tuning-free baselines and scales monotonically with the number of tutorials consumed. Together, these results demonstrate the potential of instructional videos as a scalable resource for enabling LLM agents to continually learn without parameter updates.
{\small
\bibliographystyle{unsrtnat}
\bibliography{tal}}

\newpage
\appendix

\section{Implementation and Evaluation Details}
\label{sec:implementation_evaluation_details}

\subsection{Baseline and Experimental Configuration}
\label{sec:baseline_experimental_configuration}

The existing RAG baseline cannot learn directly from raw video inputs and therefore operates on video transcripts rather than the videos themselves. We construct a subtitle corpus from the tutorial videos and segment each subtitle stream into overlapping 220-token chunks with a 50-token overlap, retaining metadata including the opening name and video title. The retrieval query concatenates the question and answer options. Retrieval is performed with a hybrid pipeline consisting of BM25 and a dense retriever (OpenAI's text-embedding-3-small) based on cosine similarity. Each retriever returns 30 candidates, whose ranked lists are fused via Reciprocal Rank Fusion (RRF). The top five fused chunks are then supplied to the answer model as evidence.

For the Video-RAG variants, we caption each event frame and pair the caption with its local transcript. The caption-retrieval variant provides the top five retrieved text units, while the frame-aware variant additionally provides the corresponding frames.

\subsection{TAL Operations and Prompts}

\paragraph{Reflective revision of $G_\phi$.}
$G_\phi$ maintains an external read--write history $H_t$ containing previously generated questions, their teaching events, and the frozen answer model's outcomes. Before each round, it reflects on $H_{t-1}$: questions answered correctly are treated as too easy and trigger harder, non-duplicate questions about the same event, while incorrectly answered questions are retained as knowledge gaps in the QA bank. The validated questions and outcomes are then written back to $H_t$ to guide the next round. Thus, reflective revision updates the generator's external state and prompt context across rounds rather than its parameters.

\begin{promptbox}{P.1 \quad Reflective revision of the question generator}
Hard rewrite mode:
- Rewrite attempt: {attempt_idx}
- The previous question was answered correctly by the validator model.
- Generate ONE new and harder MCQ on the same event/frame.
- Do not reuse the same stem, option set, or answer key.
- Prefer move-order traps, concrete tactical refutations, or subtle plan selection.
- Make distractors very plausible and close to the best move.
- Previous question: {prev_q} | Previous correct option: {prev_answer}
\end{promptbox}

\paragraph{Tutorialbook editing.}
After evaluating the current tutorialbook, TAL routes each failed QA to its implicated chapter. For each chapter, the editor examines the failed questions, correct and common-wrong answers, explanations, and evidence to identify missing or misleading knowledge. It then adds or corrects the relevant content while preserving useful existing material. The revised chapter is re-evaluated on the QA bank and committed only if exact-match accuracy strictly improves; otherwise, the previous chapter is restored.

\begin{promptbox}{P.2 \quad Stage II tutorialbook editor (Write operator)}
[system]
You are refining a chess tutorialbook chapter using failed quiz questions.
Given the current chapter and failed questions, improve instructional quality and
question traceability. Integrate missing knowledge into the most relevant existing
sections whenever possible (opening ideas, move purpose, plans, tactical motifs,
common mistakes). If needed, add new sections, but do not bloat the chapter.
[user]
Current chapter: {textbook}
Failed questions (id/options/correct/common-wrong/hint/explanation/context):
{failed_questions}
Update the chapter to cover these gaps. Return JSON only.
\end{promptbox}

\Needspace{0.36\textheight}
\paragraph{TOC retrieval.}
TAL serializes the tutorialbook tree as a compact index containing each chapter's category/subtheme/chapter path, ID, and summary. Given this index, the question, and its options, the frozen LLM selects up to three relevant chapter IDs in JSON. TAL validates the IDs and expands only those chapters into the answer context. This tree-based retrieval avoids reading the full tutorialbook or relying on embedding similarity, achieving 50.0\% Recall@3 compared with 28.2\% for dense retrieval (Table~\ref{tab:ablation_retrieval_backend}). Each added video increases the compact TOC by approximately 240 tokens, corresponding to 0.024\% of a 1M-token context window. If the TOC eventually exceeds the available context, hierarchical or multi-stage TOC navigation provides a natural extension.

\begin{promptbox}{P.3 \quad TOC retrieval}
[system]
You choose the most relevant tutorialbook chapters for one chess MCQ.
Only use the provided TOC.
Return strict JSON only: {"chapter_ids": [up to K ids], "confidence": number}.
[user]
Question: {question_text}
Options: {options_text}
Tutorialbook TOC:
  toc_1: category=... | subtheme=... | chapter=... | summary=...
  toc_2: ...
Return JSON only.
\end{promptbox}

\subsection{Judge Behavior and Verification}
A frozen LLM judge is used only as the Stage~I grounding gate. Given a candidate QA pair and its local subtitle window, it determines whether the reference answer is supported by the lecturer's explanation. A manual audit of 50 judge decisions found 100\% agreement with human checks, with all retained answers traceable to subtitle evidence. The judge does not score reported metrics: accuracy, difficulty filtering, and Stage~II acceptance use exact MCQ matching, and an edit is retained only if QA-bank accuracy improves. All 110 held-out questions were independently checked by two people.

\subsection{Construction Cost}
\label{sec:construction_cost}
At API prices of \$0.50 per million input tokens and \$3 per million output tokens, Gemini-3-Flash construction uses a measured average of 245k input tokens and 88k output tokens over 102 calls per video. This corresponds to approximately \$0.38 per video (\$0.12 input and \$0.26 output), or \$0.03 per minute for an average video duration of 12.9 minutes. Construction is a one-time cost, and the resulting tutorialbook can be reused across different answer models without reconstruction.

\subsection{Robustness to Noisy or Irrelevant Videos}
Videos are collected automatically by a video-search agent that queries topic keywords and selects the top-$k$ highly viewed matches, providing a lightweight relevance and popularity filter. As the corpus grows over 12/24/36/48 videos, Transcript-RAG test accuracy fluctuates from 42.7\% to 38.2\%, 43.6\%, and 37.3\%, whereas TAL improves from 48.2\% to 57.3\%, 61.8\%, and 64.6\% under the same video budgets. These results suggest that TAL can tolerate naturally occurring variation in source quality and relevance, although robustness to controlled corruptions remains an important direction for future work.

\section{VidLearn Details}
\label{sec:vidlearn_details}
For each of the 12 classical openings (Table~\ref{tab:vidlearn_openings}), we issue the opening name as a YouTube search query and download the top results, retaining instructional videos that (i) carry usable English subtitles, (ii) have titles topically relevant to the queried opening (verified by an LLM-based title-relevance audit), and (iii) are not duplicates within the same opening (deduplicated by video ID). After filtering, we obtain $72$ videos in total ($6$ per opening; $4$ for training and $2$ held-out for testing).  The filtered corpus spans $\sim$$14.9$ hours and $\sim$$462$K subtitle words across the 48 training videos, plus $\sim$$7.7$ hours and $\sim$$262$K words across the 24 held-out test videos ($\sim$$22.5$ hours and $\sim$$724$K words in total).

All videos in VidLearn are publicly hosted on YouTube. To respect content creators' rights and YouTube's Terms of Service, we release
only per-question metadata (video ID, timestamp, question, options, answer index, and verified rationale) along with the question-generation and validation code, sufficient for full reproduction. Raw video files will not be redistributed.

\begin{table}[ht]
\centering
\caption{The 12 classical openings covered in VidLearn.}
\label{tab:vidlearn_openings}
\begin{tabular}{lll}
\toprule
\multicolumn{3}{c}{Opening} \\
\midrule
The Caro-Kann Defense & The French Defense & The Gr\"unfeld Defense \\
The Italian Game & The King's Indian Defense & The London System \\
The Nimzo-Indian Defense & The Queen's Gambit & The Ruy Lopez \\
The Scotch Game & The Sicilian Defense & The Slav Defense \\
\bottomrule
\end{tabular}
\end{table}

\section{Human Performance}
\label{sec:human_performance}
\begin{figure}[!htbp]
\centering
\includegraphics[width=0.62\linewidth]{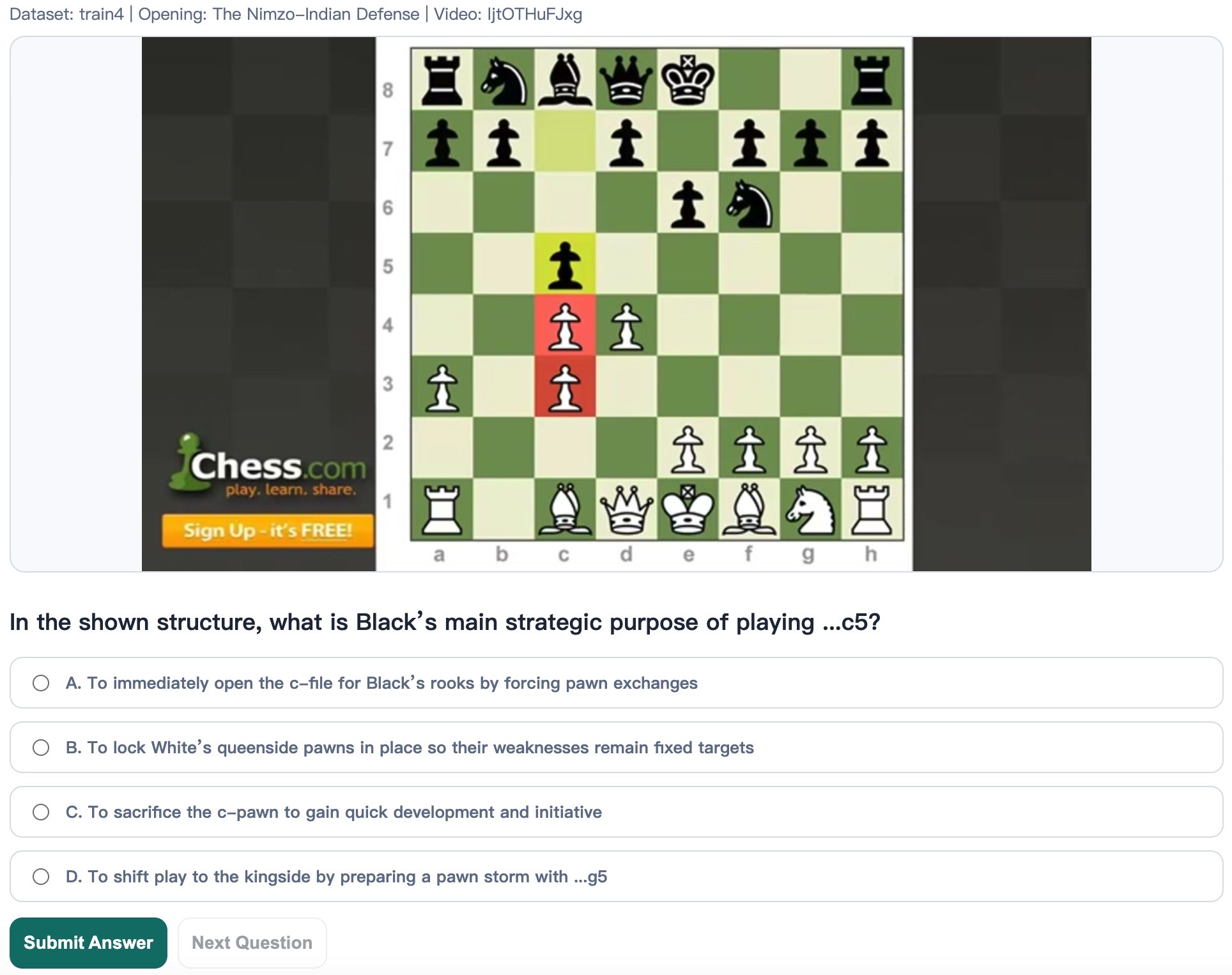}
\caption{Interactive user interface used for human evaluation.}
\label{fig:human_eval}
\end{figure}

One of the motivations behind TAL is to move closer to how humans acquire knowledge: by learning directly from demonstrations and explanations, rather than only from parameter updates or interaction traces. To provide a qualitative reference for this learning paradigm, we additionally collect human responses on the same held-out test questions and compare the resulting trend with that of TAL.
To collect human responses under the same protocol used for TAL, we built an interactive evaluation interface (Fig.~\ref{fig:human_eval}) that presents each test instance exactly as it is given to the model and enforces closed-book conditions (no engines, books, or web search). We answered the held-out test questions through this interface after watching $K_v \in \{12, 24, 36, 48\}$ training videos, with each $K_v$ condition completed in a separate session. This small-scale reference is used only to check whether the same learning-from-videos trend observed for TAL also appears for a human under the same protocol, and is not intended as a statistically representative estimate of human performance.

\begin{figure}[!htbp]
\centering
\includegraphics[width=\linewidth]{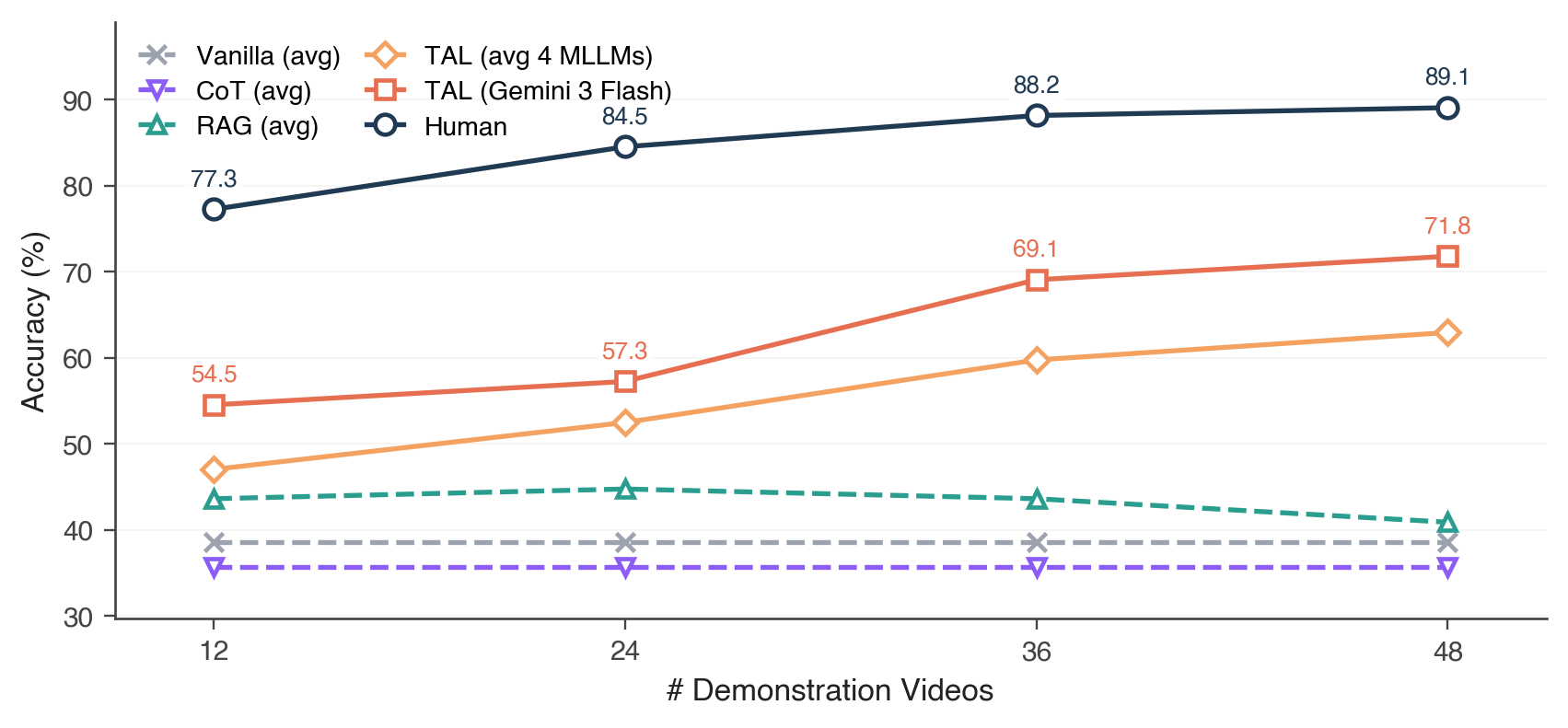}
\caption{Comparison of learning curves on the held-out test set.}
\label{fig:human_perform}
\end{figure}

Both the human and TAL (avg) curves rise with the number of demonstration videos, supporting the central premise that instructional videos provide a meaningful learning signal for both humans and TAL. The two curves, however, differ in shape in ways that are informative for TAL's design. Humans exhibit strong consolidation efficiency: by extracting a shared underlying logic across stylistically and pedagogically diverse presentations of the same opening, they acquire the relevant knowledge from only a handful of tutorials---often a single high-quality demonstration.

\begin{samepage}
This contrast highlights both a current limitation and an opportunity for TAL. The limitation is that, under the same evaluation protocol, TAL's accuracy remains below the human reference. The opportunity is that TAL's curve scales monotonically with the number of demonstration videos in the regime tested: each additional video continues to inject useful structure into tutorial memory. This suggests that TAL's ability can continue to grow as more tutorials are consumed.
\par
\end{samepage}

\FloatBarrier

\section{Train--Test Similarity Analysis}
To assess the risk of train--test leakage, we compute nearest-neighbor overlap statistics between each of the $110$ held-out test questions and two reference pools: (i) the harvested question bank $\mathcal{Q}_t^{\text{bank}}$, consisting of $174$ QA pairs generated by Stage~I from the training videos (hereafter referred to as training questions); and (ii) the set of all subtitle chunks extracted from the training videos. The subtitle pool comprises $2{,}564$ chunks, obtained from the $48$ training videos using a chunk size of $220$ tokens with a $50$-token overlap, followed by exact-text deduplication.

For each test question, we report the maximum similarity to any candidate in each reference pool. All texts are normalized by lowercasing, stripping punctuation, and collapsing repeated whitespace. TF-IDF features are extracted with unigrams and bigrams (\texttt{ngram\_range=(1,2)}); for latent semantic similarity, we project the TF-IDF vectors into a $128$-dimensional subspace via \texttt{TruncatedSVD} and compute cosine similarity after $L_2$ normalization. For each metric, we report the mean$\pm$SD, the bootstrap $95\%$ confidence interval ($3{,}000$ resamples at the question level, \texttt{seed=7}), the median, the $90$th percentile (P90), and the proportion of test questions with similarity score $\geq 0.8$, which serves as a near-duplicate indicator.

\begin{table*}[h]
\centering
\caption{Nearest-neighbor overlap between each of the $110$ test questions and the $174$ training questions (left) / $2{,}564$ training subtitle chunks (right). Lower values indicate weaker train--test similarity. No test question has a near-duplicate match ($\geq 0.8$) under any metric or pool, ruling out trivial lexical leakage.}
\label{tab:train_test_overlap_stats}
\resizebox{\textwidth}{!}{%
\begin{tabular}{lccccc|ccccc}
\toprule
 & \multicolumn{5}{c|}{\textbf{Against Train Questions (n=174)}} & \multicolumn{5}{c}{\textbf{Against Train Subtitle Chunks (n=2564)}} \\
\cmidrule(lr){2-6}\cmidrule(lr){7-11}
\textbf{Metric} & Mean$\pm$SD & 95\% CI & Median & P90 & $\geq0.8$ & Mean$\pm$SD & 95\% CI & Median & P90 & $\geq0.8$ \\
\midrule
Ngram-2 Overlap & 0.295$\pm$0.116 & [0.273, 0.317] & 0.282 & 0.452 & 0.0\% & 0.208$\pm$0.062 & [0.196, 0.219] & 0.200 & 0.286 & 0.0\% \\
Max Jaccard & 0.389$\pm$0.073 & [0.377, 0.403] & 0.393 & 0.468 & 0.0\% & 0.184$\pm$0.029 & [0.179, 0.190] & 0.185 & 0.219 & 0.0\% \\
TF-IDF Cosine & 0.210$\pm$0.065 & [0.198, 0.223] & 0.207 & 0.288 & 0.0\% & 0.100$\pm$0.024 & [0.095, 0.104] & 0.097 & 0.131 & 0.0\% \\
LSA Cosine (128d) & 0.419$\pm$0.115 & [0.397, 0.442] & 0.407 & 0.574 & 0.0\% & 0.426$\pm$0.079 & [0.411, 0.440] & 0.423 & 0.513 & 0.0\% \\
\bottomrule
\end{tabular}%
}
\end{table*}

Across all four similarity metrics and both reference pools, no test question exhibits a near-duplicate match ($\geq 0.8$) with any training question or subtitle chunk (Table~\ref{tab:train_test_overlap_stats}). Lexical overlap remains consistently low under bigram overlap, Jaccard similarity, and TF-IDF cosine, ruling out the possibility that test performance is driven by direct reuse of training questions or subtitle text. LSA cosine indicates moderate topic-level relatedness---expected, since both splits draw from the same $12$ openings---but even under this latent semantic view, no near-duplicate overlap is observed.
\section{Online Self-Evolution}
\label{sec:online_self_evolution}
The TAL tutorial memory framework supports not only offline construction from a fixed video collection, but also on-demand expansion. When the  online-learning mode is enabled, the agent inspects the Table of Contents (TOC) of the current memory upon receiving a target topic  (e.g., the name of an opening); if no matching entry is found in the TOC, the topic is treated as a knowledge gap, and the agent generates a set of search queries to retrieve relevant instructional videos from external sources such as YouTube. The retrieved videos are then processed through the same Stage I and Stage II refinement pipeline used during initial construction. This can be viewed as a lightweight closed loop---keyword $\rightarrow$ gap detection $\rightarrow$ video discovery $\rightarrow$ memory update---which lets the memory grow without any parameter update.

We exercise this mode end-to-end in our VidLearn experiments: the agent is given only the names of the $12$ openings, without any pre-curated
corpus or human-filtered video list, and constructs its entire training corpus through the TOC-based gap-detection mechanism alone.  TAL therefore does not depend on manually curated sources; the framework itself can extend its knowledge base in open environments, leaving room for deployment to broader and faster-changing domains.

\end{document}